\documentclass[letterpaper,10pt,conference]{ieeeconf}

\IEEEoverridecommandlockouts
\usepackage{amsmath}
\usepackage{amssymb}
\usepackage{booktabs}
\usepackage{graphicx}
\usepackage{gensymb}
\usepackage{xcolor}
\usepackage{xspace}
\usepackage{hyperref}
\usepackage{cite}
\usepackage{array}
\usepackage{soul}
\usepackage{multirow}
\usepackage{caption}

\usepackage{cuted}    
\usepackage{capt-of}
\definecolor{todocolor}{RGB}{170,35,35}

\newcommand{\uppergoal}{\mathbf{g}^u_t}
\newcommand{\lowergoal}{\mathbf{g}^l_t}
\newcommand{\proprio}{\mathbf{s}^p_t}

\newcommand{\upperaction}{\mathbf{a}^u_t}
\newcommand{\loweraction}{\mathbf{a}^l_t}

\newcommand{\SE}[1]{\mathrm{SE}(#1)}

\newcommand{\method}{\textsc{TapNav}\xspace}

\title{\LARGE \bf
\method: Humanoid Navigation through Tactile Active Perception
}

\author{
Huaze Liu$^{*1,2}$,
Zhenyu Wu$^{*1}$,
Jaehwi Jang$^{*1}$,
Junjie Sheng$^{1}$,
Andrew Collins$^{1}$,
Aaron Xie$^{3}$,
Zhaoyuan Gu$^{1}$,\\
Kaijie Zhu$^{1}$,
Ding Jiang$^{1}$,
Kevin Cai$^{1}$,
Ye Zhao$^{1}$\\[2pt]
$^{*}$Equal Contribution
\qquad
$^{1}$Georgia Institute of Technology
\qquad
$^{2}$Harvey Mudd College
\qquad
$^{3}$Walton High School
}

\begin{document}

\maketitle
\thispagestyle{empty}
\pagestyle{empty}

\begin{strip}
\centering
\includegraphics[width=1.0\linewidth]{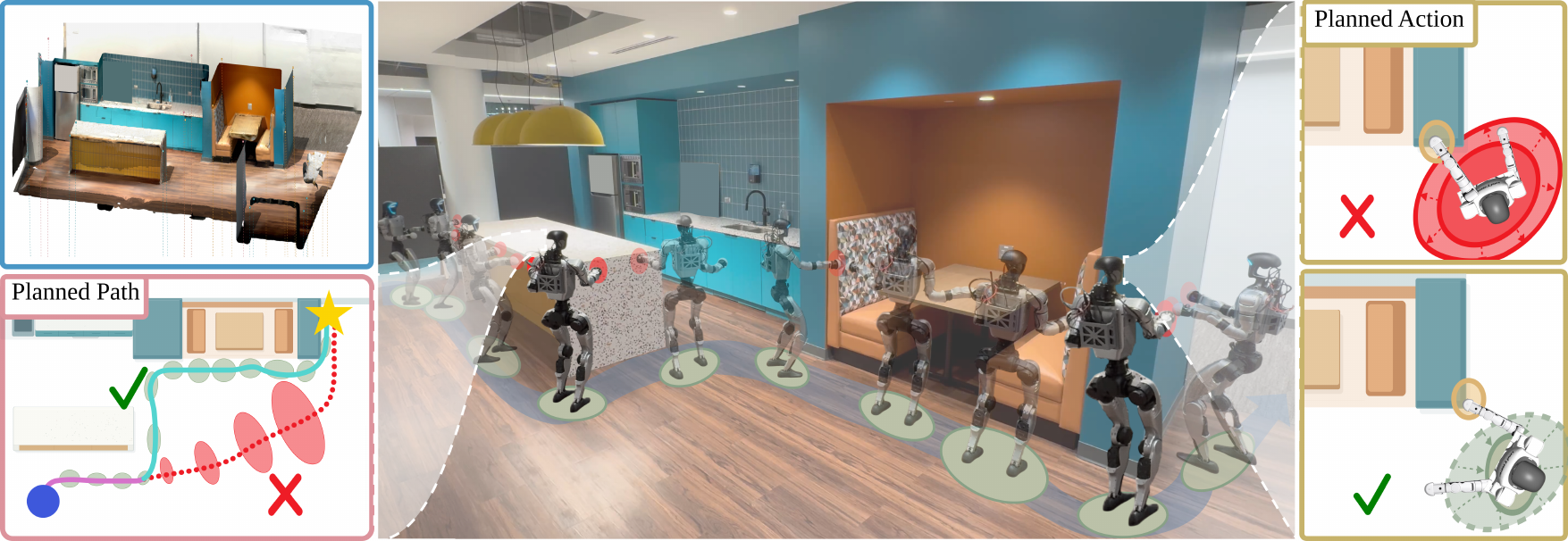}
\captionof{figure}{
\method enables vision-denied humanoid navigation through tactile active perception.
The humanoid actively makes contact with the surrounding environment at informative locations,
using contact observations to reduce localization uncertainty and achieve reliable navigation.
}
\label{fig:teaser}
\end{strip}

\begin{abstract}
Navigation in vision-denied environments is challenging for humanoid robots because proprioceptive odometry drifts and localization uncertainty accumulates rapidly. We present \method, a tactile active-perception framework that enables humanoid navigation toward a goal by actively probing surrounding structures without relying on vision. \method maintains a pose belief from odometry, IMU, and tactile contact observations, and couples uncertainty-aware global route planning with information-gain-driven local probing. The global planner searches for routes that keep predicted localization uncertainty bounded by exploiting opportunities for tactile correction, while the local planner selects probe actions that maximize expected information gain. A whole-body controller coordinates the humanoid's locomotion and end-effector contact to execute the planned navigation and probe motions. We evaluate \method in simulation and on a Unitree G1 across different floor plans and obstacle geometries. \method achieves lower state estimation error and a higher task completion rate than baselines. These results demonstrate that actively planning physical interactions with the environment can provide localization cues for reliable humanoid navigation without vision. More results are available at \href{https://tapnav.github.io/TAPNAV-Project/}{\method project page}.

\end{abstract}

\section{Introduction}

Humanoid robots hold great promise for bringing intelligent physical assistance to human-centered environments \cite{gu2026humanoid}. To succeed in indoor environments, humanoid robots require not only robust locomotion but also reliable localization and navigation under safety-critical and perceptually degraded conditions. This capability is particularly important in hazardous or inaccessible scenarios such as fire emergency response, industrial gas-leak inspection, and engine-room operation with degraded lighting.

Recent progress in humanoid perception and navigation has largely relied on RGB-D cameras and LiDAR for environment sensing and state estimation \cite{zhu2026emobipednav, muenprasitivej2024bipedal}. These sensing modalities, however, can become unreliable in visually degraded environments, 
compromising both localization and navigation capabilities. In contrast, humans can compensate for limited vision through haptic interactions, which offer an alternative source of geometric information when visual observations are unreliable.

Motivated by how people navigate with haptic feedback, this study equips a humanoid robot with tactile sensors for contact-based navigation. Traditionally, robotic tactile sensing has been widely researched in fine-grained tasks such as identifying grasped objects, estimating in-hand pose, or reconstructing local shape~\cite{han2024learning, Merwe2026RALPose}. Extending tactile sensing to humanoid navigation requires selecting informative contacts while maintaining whole-body balance. Sparse tactile observations must then be integrated with motion estimates to support localization throughout navigation.

We introduce \method: Humanoid \textbf{Nav}igation through \textbf{T}actile \textbf{A}ctive \textbf{P}erception, as shown in Fig.~\ref{fig:teaser}. \method is an integrated framework that guides a humanoid to a goal using tactile sensing without relying on cameras or LiDAR. To the best of our knowledge, \method is the first system to use actively planned tactile probing, rather than reactive or incidental contact, to navigate a humanoid through an office-scale indoor environment without vision.

We evaluate \method in simulation and on a Unitree G1 humanoid, covering indoor layouts with long hallways, corners, pillars, and free-standing islands.
Our contributions are summarized as follows:
\begin{itemize}
    \item A belief-space tactile navigation planner that balances route progress with localization uncertainty by exploiting opportunities for tactile correction.
    \item An active tactile probing framework that selects contact actions based on their expected information gain, accounting for both contact and contact-free outcomes to actively reduce robot pose uncertainty.
    \item An integrated vision-denied humanoid navigation system that couples belief-space planning and active tactile perception with reinforcement learning-based whole-body locomotion control, validated in both simulation and real-world hardware experiments in office-scale environments.

\end{itemize}

\section{Related Work}

\subsection{Tactile Localization and Active Sensing}

Tactile sensing has been widely applied to object-centric state estimation. Learned tactile representations and probabilistic estimation account for ambiguous contact observations, while additional contacts and sequential measurements constrain the set of plausible poses~\cite{pmlr-v205-suresh23a,Bauza_2023}. Joint pose and shape estimation has also been studied through visuo-tactile neural reconstruction and tactile-only geometric modeling, allowing object geometry to be estimated alongside its
motion~\cite{gordon2026active}.
Active sensing couples these estimation problems with the selection of subsequent interactions. Learned exploration policies use tactile feedback to select motions that improve surface coverage or object recognition~\cite{10611667,Xu_2023}. Information-based approaches evaluate candidate interactions through expected information gain or contact-aware information, selecting informative motions for estimating object pose, geometry, or physical parameters~\cite{gordon2026active,SathyanarayanH-RSS-25}. On the other hand, contact sensing has also been used for robot localization, where geometric and terrain information from gait-induced contacts is incorporated without active action selection~\cite{Buchanan2021}.

\subsection{Belief-Space Planning and Active Localization}

Belief-space planning explicitly reasons about state uncertainty when selecting navigation actions, accounting for how motion and future observations affect the robot's belief~\cite{Puthumanaillam_2026}.
Active localization further couples navigation with sensing decisions, selecting informative viewpoints or deciding when to acquire observations to improve pose estimation~\cite{10.1007/978-3-031-73016-0_12,Shek_2025}.
These approaches, however, primarily assume that localization information can be obtained through remote perceptual observations. In contrast, \method enables active localization in vision-denied settings by planning motions that leverage contact observations to maintain an accurate robot-state belief.

\subsection{Contact-Rich Humanoid Loco-Manipulation}

\begin{figure*}[t!]
    \centering
    \includegraphics[width=0.99\linewidth]{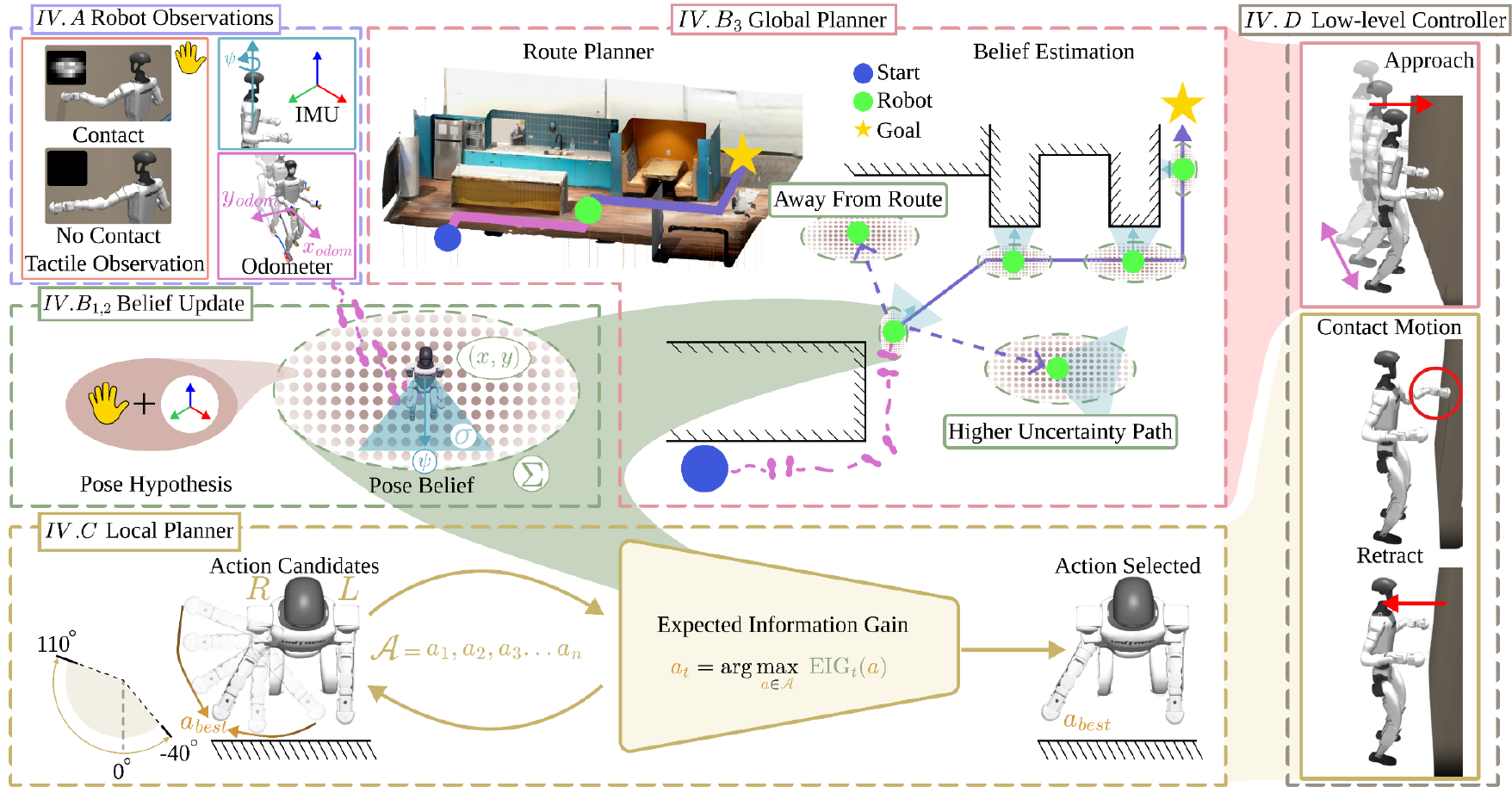}
    \caption{Overview of the \method framework.
     Odometry, IMU, and tactile observations together update the robot's $\SE{2}$ pose belief. The global planner computes an uncertainty-bounded route with predicted tactile corrections, the local planner selects informative probes based on expected information gain, and the low-level controller executes locomotion and probing.}
    \label{main_figure}
    \vspace{-0.15in}
\end{figure*}

Recent advances in humanoid loco-manipulation have enabled whole-body coordination of locomotion and manipulation for tasks such as object transport and door opening~\cite{gu2026refinedp,liu2024opt2skill}, dynamic catching~\cite{shukla2026catch}, payload transport~\cite{huang2026steadytray,jang2025seecstableendeffectorcontrol}, and force-regulated interaction~\cite{jang2026wtumitactilebasedwholebodymanipulation}. These methods demonstrate increasingly capable coordination between lower-body locomotion and upper-body manipulation, including maintaining stability under interaction forces. Unlike conventional loco-manipulation, where contact typically serves the manipulation objective, \method uses contact as a sensing action: the planner dynamically selects when and where the robot should probe to reduce localization uncertainty. 


\section{System Overview}
\label{sec:problem}
We formulate the problem as a vision-denied humanoid navigation task on a known and static map $M$. 
The robot starts from a known initial state $\mathbf{x}_0$, its state at time $t$ is represented by $\mathbf{x}_t=(p^x_t,p^y_t,\psi_t)\in\SE{2}$ . We localize the robot using odometry, an IMU-based heading estimate, and tactile contact with mapped surface. The goal is to reach a target region $\mathcal{X}_g$ while balancing route progress with active tactile probing to control localization uncertainty.

During navigation, the system coordinates a global route planner,
a local tactile action planner, and a pose estimator, as shown in
Fig.~\ref{main_figure}. At timestep $t$, given the map $M$ and the current state belief $b_t$, the global planner computes a collision-free route to
$\mathcal X_g$ that keeps
the predicted pose uncertainty bounded. The route is represented by a sequence of subgoals, each
specifying a desired displacement $\Delta\mathbf x_t$ from the preceding subgoal. 

At each subgoal, the local planner uses the latest belief to select a tactile probe that maximizes the expected information
gain (EIG) about the robot pose. A probe moves the end-effector $j$ along a straight trajectory in a direction $\alpha$ relative to the torso heading until contact is detected or a travel limit is reached. Each probe action $a$ is defined as $a = (j,\alpha) \in \mathcal{A}$ where $j\in \{L,R\}$ specifies the left or right end-effector, $\mathcal{A} = \{L,R\} \times \mathcal{A}_{\alpha}$, and $\mathcal A_\alpha$ is a uniformly discretized set of probing directions in $[-40^\circ,110^\circ]$, subject to the arm's kinematic feasibility. For both end-effectors, $\alpha=0$ points straight forward and $\alpha >0$ points outward.

The probe action $a_t$ selected by the local planner produces a tactile
observation $\mathbf{o}_t^{\mathrm{tac}} = (\mathbf s_t,c_t)$, where $\mathbf s_t \in \mathbb{R}^2$ is the end-effector stopping position in the body frame, and $c_t \in \{0, 1\}$ is the contact outcome. $c_t=1$ indicates contact and $c_t=0$
indicates a contact-free event. The estimator maintains a pose belief $b_t(\mathbf{x}_t)=p(\mathbf{x}_t\mid \mathbf{u}_{1:t},\mathbf{o}_{1:t},M)$, where $\mathbf{u}_t =(\Delta p^x_t,\Delta p^y_t,\Delta\psi_t)\in\SE{2}$ is the body-frame odometry increment $\mathbf x_t$ from step $t-1$ to $t$, and $\mathbf{o}_t=(\mathbf{o}_t^{\text{tac}},\tilde{\psi}_t)$
collects the observations, where $\tilde{\psi}_t$ denotes the IMU heading measurement. Finally, a low-level controller executes the displacement $\Delta \mathbf{x}_t$ and the probing action $a_t$. In summary, \method actively uses touch to reduce localization uncertainty, enabling reliable goal-directed navigation without vision.

\section{Method}
\label{sec:method}

\subsection{Tactile Sensing and Contact Processing}
\label{sec:hardware}
\begin{figure}[h]
    \centering
    \vspace{-0.07in}
    \includegraphics[width=1.0\columnwidth]{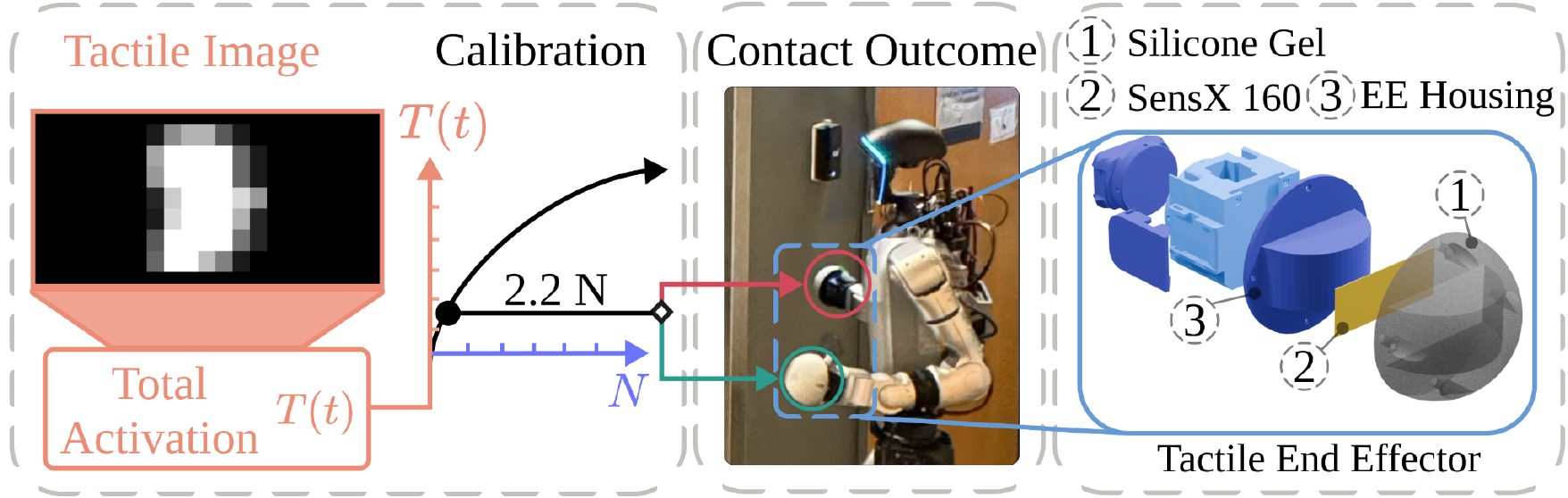}
    \caption{
     Tactile end-effector design and contact-processing pipeline. The tactile image is aggregated into a total activation $T(t)$, which is calibrated to a normal force and converted to a binary contact outcome using a force threshold. 
    }
    \label{fig:hardware_design}
    \vspace{-0.1in}
\end{figure}

The Unitree G1 is equipped with custom tactile end-effectors to detect contact with the environment, as shown in Fig.~\ref{fig:hardware_design}. Each end-effector integrates a TouchTronix SensX 160 tactile sensor covered by a silicone pad made from Ecoflex 00-50 \cite{sensx160}. The SensX 160 consists of a $20\times8$ taxel array with an active sensing area of $20\times50$ mm and a force range of up to 100 N. The sensor is mounted on a curved, rigid housing to enable contact sensing across a range of approach angles.

The tactile image is first offset-corrected and filtered to suppress sensor 
noise. We define the total activation $T(t)$ as the sum of the filtered 
taxel activations at time $t$. The total activation is then converted to an 
estimated normal force $\tilde{f}(t)$ using the inverse exponential response 
model introduced in WT-UMI 
\cite{jang2026wtumitactilebasedwholebodymanipulation}. Finally, the estimated 
normal force is thresholded to produce a binary contact outcome, which is provided to the planner as a tactile observation.

\subsection{Belief-Space Tactile Navigation Planning}
\label{sec:planner}

\subsubsection{Pose belief}
At each time step $t$, we maintain a pose belief $b_t$ over the robot's current pose $\mathbf{x}_t$ using a particle filter, which is then acted on by the global and local planners. During locomotion, odometry propagates the pose hypotheses, while tactile and IMU observations correct the accumulated drift. The belief is represented by $N_p$ weighted particles $\{{\mathbf x_t^{(1)}, \dots, \mathbf x_t^{(i)}, \dots, \mathbf x_t^{(N_p)}\}}$,
\begin{equation}\nonumber
b_t(\mathbf{x}_t)
=
p(\mathbf{x}_t\mid \mathbf{u}_{1:t}, \mathbf{o}_{1:t},M)
\approx
\sum_{i=1}^{N_p}
w_t^{(i)}\,
\delta\!\left(\mathbf{x}_t-\mathbf{x}_t^{(i)}\right),
\label{eq:particle_belief}
\end{equation}
where $\mathbf{x}_t^{(i)} = ({p^x_t}^{(i)}, {p^y_t}^{(i)}, {\psi^{(i)}_t}) \in \SE{2}$ is the $i$-th pose hypothesis, and $w_t^{(i)}$ is its normalized weight with $\sum^{N_p}_{i=1} w_t^{(i)}=1$. The particles are initialized around the known initial pose. For computational efficiency, we approximate the belief as a Gaussian $b_t(\mathbf{x}_t) \sim \mathcal{N}(\boldsymbol{\mu}_t, \boldsymbol{\Sigma}_t)$ whose mean $\boldsymbol{\mu}_t=(\mu^x_t,\mu^y_t,\mu^\psi_t) \in \mathbb{R}^3$ and covariance $\boldsymbol{\Sigma}_t\in\mathbb{R}^{3\times3}$ are computed from the weighted particles.

Given the body-frame increment $\mathbf u_t$ from odometry, each particle $i$ is propagated by composing $\mathbf x_{t-1}^{(i)}$ with $\mathbf u_t$ perturbed by Gaussian noise: 
\begin{equation}
\mathbf x_t^{(i)}=\mathbf x_{t-1}^{(i)}\oplus\big(\mathbf u_t+\boldsymbol\varepsilon_t^{(i)}\big),\qquad \boldsymbol\varepsilon_t^{(i)}\sim\mathcal N\big(\mathbf 0,Q(\mathbf u_t)\big),
\label{eq:motion_model}
\end{equation}
where $\oplus$ denotes SE(2) pose composition and $Q(\mathbf u_t)$ models odometry noise. After propagation, each particle is reweighted using tactile and heading likelihoods based on the tactile probe and measured IMU heading as
\begin{equation}%
\begin{aligned}
w_t^{(i)}
&\propto
w_{t-1}^{(i)}\,
p\!\left(\mathbf{o}_t\mid \mathbf{x}_t^{(i)},M\right) \\
&= w_{t-1}^{(i)}\, p(\mathbf{o}_t^{\text{tac}}\mid\mathbf{x}_t^{(i)},M)\,p(\tilde\psi_t\mid\psi_t^{(i)}).
\end{aligned}
\label{eq:pf_update}
\end{equation}

The tactile likelihood $p(\mathbf{o}_t^{\text{tac}}\mid\mathbf{x}_t^{(i)},M)$, described in Sec.~\ref{subsec:tactile-liklihood}, evaluates the consistency between the probe observation and the known map. The heading likelihood $p(\tilde\psi_t\mid\psi_t^{(i)})$ is a wrapped Gaussian likelihood for $\tilde\psi_t$ with mean $\psi_t^{(i)}$ and variance $\sigma_{\psi,t}^2=\sigma_{\psi,t-1}^2+\big[Q(\mathbf u_t)\big]_{\psi\psi}$, which accumulates noise over time.
The particle set is resampled when the effective sample size $N_{\text{eff}}=1/\sum_i (w_t^{(i)})^2$ falls below $0.5N_p$, with small Gaussian roughening \cite{gordon1993novel}.
\begin{figure}[t]
    \centering
    \includegraphics[width=\columnwidth]{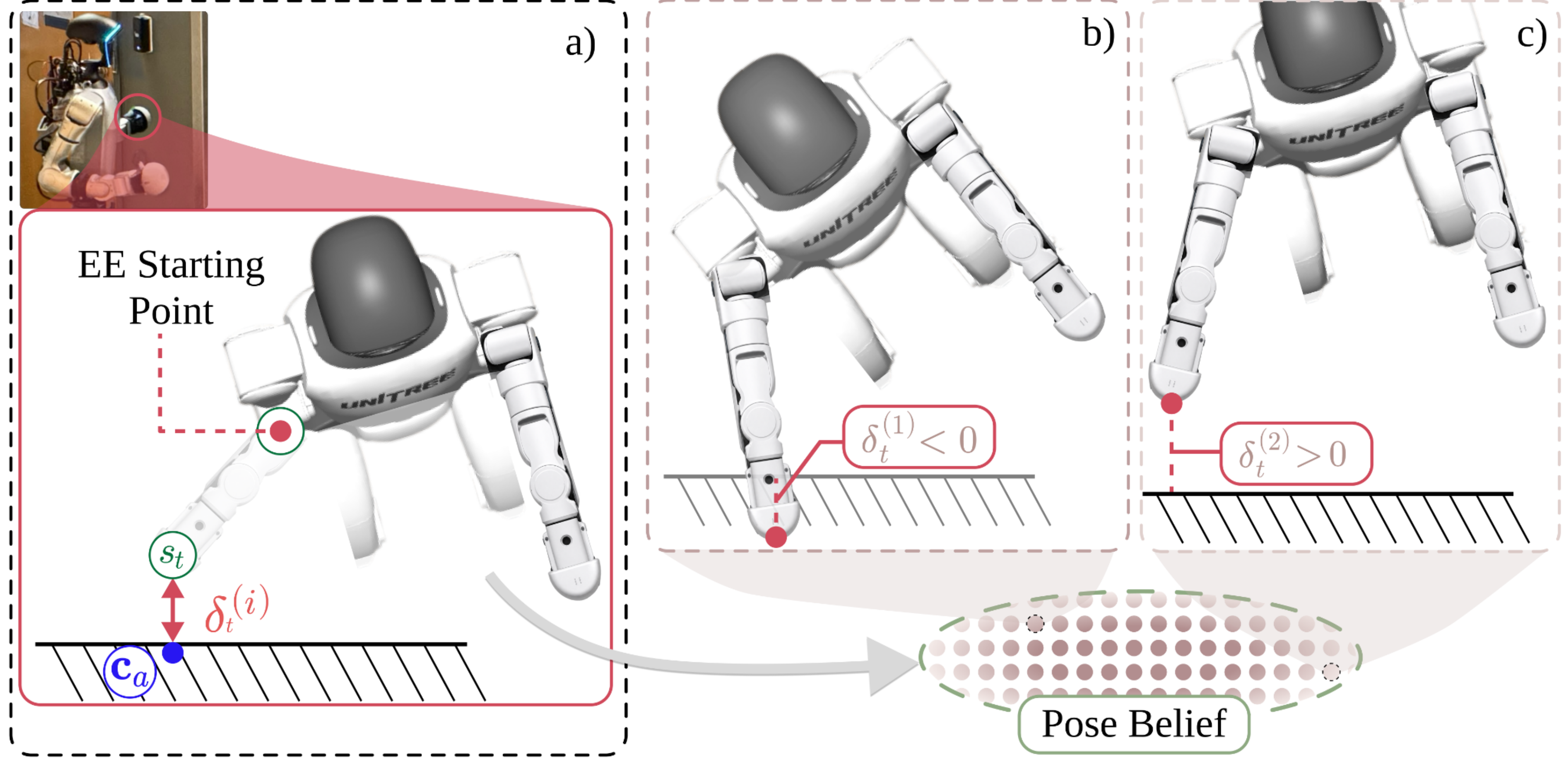}
    \caption{
    Tactile likelihood. (a) The residual $\delta^{(i)}_t$ under hypothesis $i$ and the nearest point on the map surface $\mathbf c_t^a$ defined in Sec.~\ref{sec:eig_planner}; (b) Under hypothesis $1$, the end-effector would penetrate the surface ($\delta_t^{(1)}<0$); (c) Under hypothesis $2$, it would not reach the surface ($\delta_t^{(2)}>0$).}
    \label{fig:contact_rep}
    \vspace{-0.15in}
\end{figure}

\subsubsection{Tactile likelihood} 
\label{subsec:tactile-liklihood}

A tactile probe provides geometric information about the robot pose by relating the observed end-effector stopping position to the known map. 

For each particle $\mathbf x_t^{(i)}$, we transform the observed body-frame stopping position $\mathbf s_t$ into the map frame. We denote the transformed stopping point's signed normal
distance to the nearest map surface by $\delta_t^{(i)}$.  $\delta_t^{(i)} = 0$ indicates that the stopping point lies on the surface, $\delta_t^{(i)} < 0$ indicates predicted penetration beyond the surface, and $\delta_t^{(i)} > 0$ indicates that the probe stops before reaching the surface, as illustrated in Fig.~\ref{fig:contact_rep}. 

For a contact observation ($c_t=1$), a consistent pose hypothesis should place the stopping point on the surface, and therefore hypotheses are penalized according to $|\delta_t^{(i)}|$. For a contact-free observation ($c_t=0$), a hypothesis is inconsistent only when $\delta_t^{(i)} < 0$, since it predicts that the end-effector should have encountered the surface before reaching its observed stopping position. We therefore define the tactile likelihood as
\begin{equation}
p\!\left(o_t^{\mathrm{tac}}\mid\mathbf x_t^{(i)},M\right)
\propto
\begin{cases}
\displaystyle
\exp\!\left(-\frac{(\delta_t^{(i)})^2}{2\sigma_c^2}\right),
& c_t=1,\\[1em]
\displaystyle
\exp\!\left(
-\frac{\big[-\delta_t^{(i)}\big]_+^2}{2\sigma_c^2}
\right),
& c_t=0,
\end{cases}
\label{eq:tactile_likelihood}
\end{equation}
where $[\,\cdot\,]_+=\max(\,\cdot\,,0)$ and $\sigma_c$ is the measurement noise for contact. Thus, contact observations favor pose hypotheses that place the stopping point on a map surface, whereas contact-free observations downweight hypotheses that predict a surface should have been encountered during the probe. The resulting likelihood is used in \eqref{eq:pf_update} to update and normalize the particle weights.

\subsubsection{Uncertainty-bounded global planner}
\label{sec:global-planner}
In vision-denied navigation, the geometrically shortest route may not be the most reliable route, as localization uncertainty can grow rapidly when the robot traverses regions without opportunities for tactile correction. The global planner searches for a short route subject to predicted localization uncertainty bounds, favoring contact with map surfaces when tactile probing correction is needed.

We impose position and heading uncertainty bounds $(\bar\sigma_p,\bar\sigma_\psi)$ along a route and limit route displacements to $|\Delta\mathbf x_t|\leq 0.4$~m.
Let $\Delta \mathbf{x}_t^{xy}$ denote the position entries of a displacement $\Delta \mathbf{x}_t$, $\Delta \mathbf{x}_t^{\psi}$ the yaw entry, $\mathbf{\Sigma}^{xy}_t$ the position block of the covariance $\mathbf\Sigma_t$, and $\Sigma^{\psi\psi}_t$ the yaw variance. The bounds $(\bar\sigma_p,\bar\sigma_\psi)$ are imposed at every displacement $\Delta\mathbf{x}_t$:
\begin{equation}
\begin{aligned}
    \sqrt{
        (\Delta \mathbf{x}_t^{xy})^\top
        \mathbf{\Sigma}_t^{xy}
        \Delta \mathbf{x}_t^{xy}
    }
    &\leq \bar{\sigma}_p,
    \;
    \sqrt{
        (\Delta \mathbf{x}_t^\psi)^2
        \Sigma^{\psi\psi}_t
    }
    \leq \bar{\sigma}_\psi.
\end{aligned}
\label{eq:global_problem}
\end{equation}

To evaluate \eqref{eq:global_problem}  during route search, the planner predicts how the pose belief evolves along each candidate route. Propagating the full particle set for every candidate is computationally intractable. For each candidate displacement \(\Delta\mathbf x_t\), our planner first propagates \((\boldsymbol\mu_t,\boldsymbol\Sigma_t)\) through \eqref{eq:motion_model}, accounting for the uncertainty accumulated during locomotion. We also chose the robot heading to face the nearest reachable map surface to allow tactile probing after a displacement.

The planner then predicts the effect of tactile probing using a simplified observation model. For global route evaluation, when a surface is within the probing range of a predicted mean pose, it assumes contact with a local plane of normal $\mathbf n_t$ and applies a Kalman update that reduces the covariance \(\boldsymbol\Sigma_t\) along \(\mathbf n_t\) according to the contact likelihood in \eqref{eq:tactile_likelihood}. Otherwise, the propagated covariance is retained without a tactile update. This approximation captures how proximity to environmental structure can constrain localization uncertainty without simulating actual probing during route search.

A constrained A\(^*\) search applies this belief propagation, predicts the tactile update at each candidate displacement, and prunes candidates that violate \eqref{eq:global_problem}. It returns the shortest remaining route as a sequence of displacement subgoals. The local planner in Sec.~\ref{sec:eig_planner} then selects a probe action at each subgoal. During execution, actual observations update the belief, and the route is replanned when the estimated pose deviates from it beyond a threshold.

\subsection{Expected-Information-Gain-Driven Action Selection}
\label{sec:eig_planner}

At each subgoal along the global route, the robot probes the environment with its tactile end-effectors to bound the localization uncertainty. We determine the probe action based on the probabilities and information gains of the possible outcomes of each candidate action to maximize the expected reduction in uncertainty. Under the belief $b_t\sim \mathcal{N}(\boldsymbol{\mu}_t, \boldsymbol{\Sigma}_t)$, each candidate action $a\in\mathcal A$ can result in one of two possible outcomes: either a contact or a contact-free observation.

\subsubsection{Contact geometry of an action candidate}

For each candidate action $a$, we first compute the hypothetical end-effector position at the travel limit of its straight-line probing trajectory under the current mean pose $\boldsymbol{\mu}_t=(\mu^x_t,\mu^y_t,\mu^\psi_t)$. We find the map-surface point closest to this predicted end-effector position and express it in the body frame as $\mathbf c_t^a$, as illustrated in Fig.~\ref{fig:contact_rep}(a). The surface around \(\mathbf c_t^a\) is locally approximated as a plane with outward normal \(\mathbf n_t^a\in\mathbb{R}^2\) in the map frame $M$.

To quantify how the predicted probe-to-surface geometry affects pose uncertainty, we linearize the signed normal distance to this local plane around $\boldsymbol\mu_t$. Its Jacobian with respect to the robot pose is $H_t^a= [\,
  (\mathbf n_t^a)^\top,\;
  (\mathbf n_t^a)^\top R(\mu_t^\psi+\tfrac{\pi}{2})\,\mathbf
  c_t^a
  \,]\in\mathbb R^{1\times3},$
where $R(\mu_t^\psi)$ denotes the rotation from the body frame to the map frame at the mean heading $\mu_t^\psi$. Thus, $H_t^a$ maps errors in robot pose to changes in the signed normal distance.

\subsubsection{Contact probability}
\label{contact_probability}
\begin{figure}[t]
\centering
\includegraphics[width=\columnwidth]{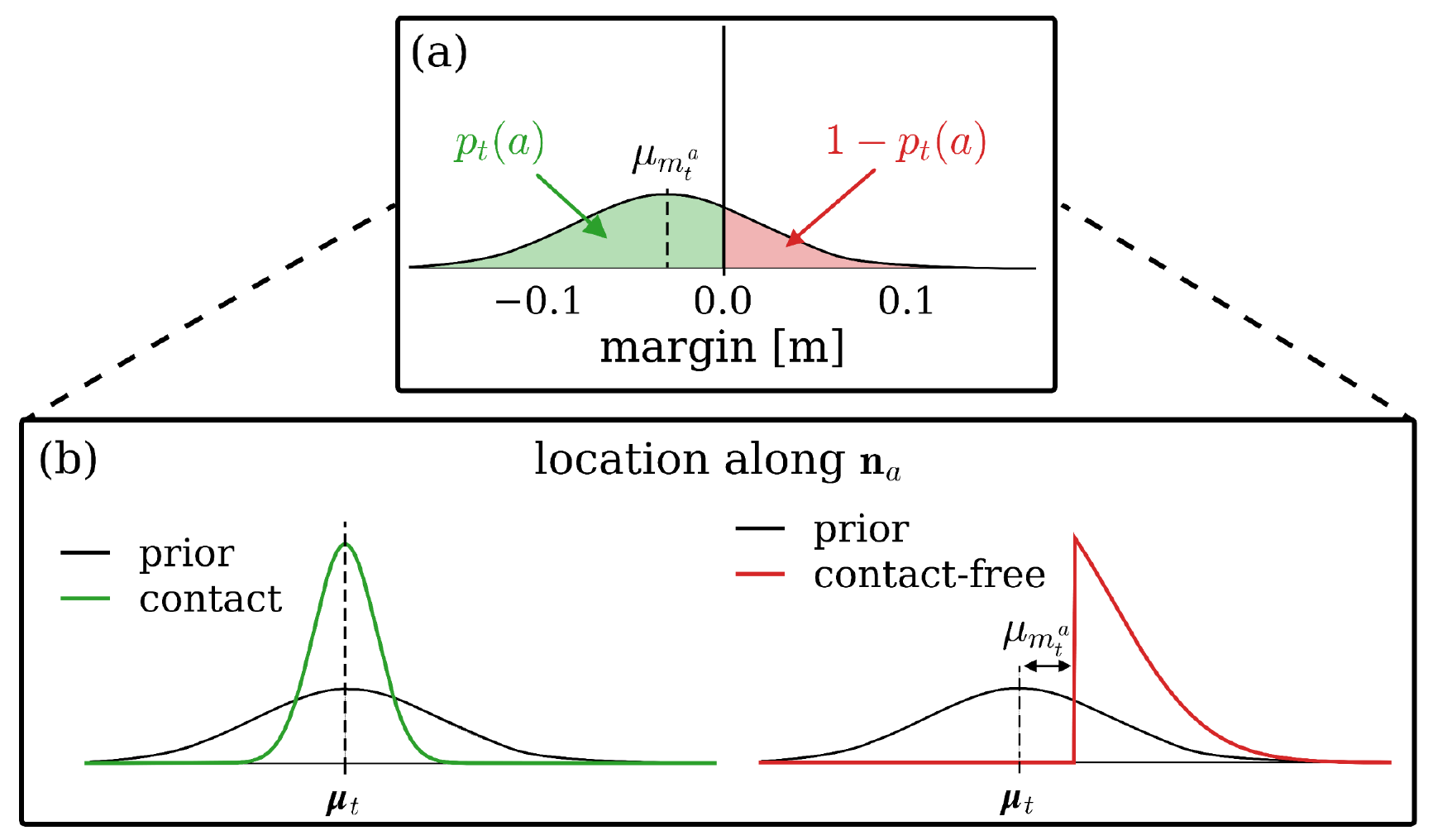}
\caption{Information gain of one probe action. (a) The margin uncertainty, modeled as $\tilde m_{t}^{a} \sim \mathcal{N}(\mu_{m_{t}^{a}}, \sigma^2_{m_t^a})$. (b) Pose belief along $\mathbf n_t^a$ before the probe, after a contact, and after a miss (prior truncated at $(\mathbf n_t^a)^\top \boldsymbol\mu_t^{xy} - \mu_{m_{t}^{a}}$).}
\label{fig:ig_hit_miss}
\end{figure}

At the current mean pose $\boldsymbol\mu_t$, we model the margin $\tilde m_t^a$, defined as the signed normal distance from the end-effector position at its travel limit to the local surface at $\mathbf c_t^a$, as a Gaussian random variable. Under this definition, $\tilde m_{t}^{a} < 0$ indicates that the surface lies within the probe's reach. $\tilde m_{t}^{a}$ is subject to uncertainty arising from both the pose uncertainty $\boldsymbol\Sigma_t$ and the control noise, modeled as $\tilde m_{t}^{a} \sim \mathcal{N}(\mu_{m_{t}^{a}}, \sigma^2_{m_t^a})$ with

\begin{equation}
  \sigma^2_{m_t^a}
  =
  H_t^a\boldsymbol\Sigma_t H_t^{a^\top}
  +
  \sigma_{\mathrm{control}}^2,
  \label{eq:margin_uncertainty}
  \end{equation}
where $H_t^a\boldsymbol{\Sigma}_tH_t^{a^\top}$ is the margin variance of the end-effector's normal coordinate induced by the pose uncertainty $\boldsymbol{\Sigma}_t$ and \(\sigma_{\mathrm{control}}^2\) accounts for controller noise. We can model the probability of contact as $p_t(a) = p(\tilde m_{t}^{a} < 0) =\Phi\!\left(-\mu_{m_{t}^{a}}/\sigma_{m_t^a}\right)$, where $\Phi$ is the standard normal cumulative distribution function, as shown in Fig.~\ref{fig:ig_hit_miss}(a).


\subsubsection{Expected information gain}
Next, we quantify the reduction in pose uncertainty for each possible probe outcome. For a contact observation, the signed normal distance to the surface provides a Gaussian measurement with Jacobian \(H_t^a\) and contact measurement noise variance \(\sigma_c^2\). The resulting information gain is
$\mathrm{IG}_t^{\mathrm{c}}(a)
=
\frac{1}{2}
\log\!\left(
1+
H_t^a\boldsymbol{\Sigma}_tH_t^{a^\top}/\sigma_c^2
\right)$~\cite{cover2006elements}.

A contact-free observation establishes that the probe did not encounter the surface, corresponding to the inequality $\tilde m_{t}^{a} > 0$. This truncates the margin distribution in \eqref{eq:margin_uncertainty} at zero. Its information gain is $\mathrm{IG}_t^{\mathrm{f}}(a)
=
-\frac{1}{2}
\log\!\big[
1-f_t^a(1-\varrho_t^a)
\big]$, 
where $f_t^a = H_t^a\boldsymbol\Sigma_t H_t^{a\top} / \sigma_{m_t^{a}}^2$ is the fraction of margin variance attributable to pose uncertainty and $\varrho_t^a$ is the variance ratio of the truncated normal distribution relative to its prior. Fig.~\ref{fig:ig_hit_miss}(b) shows the resulting belief reduction under contact and contact-free outcomes.

\begin{figure}[t]
    \centering
    \includegraphics[width=\linewidth]{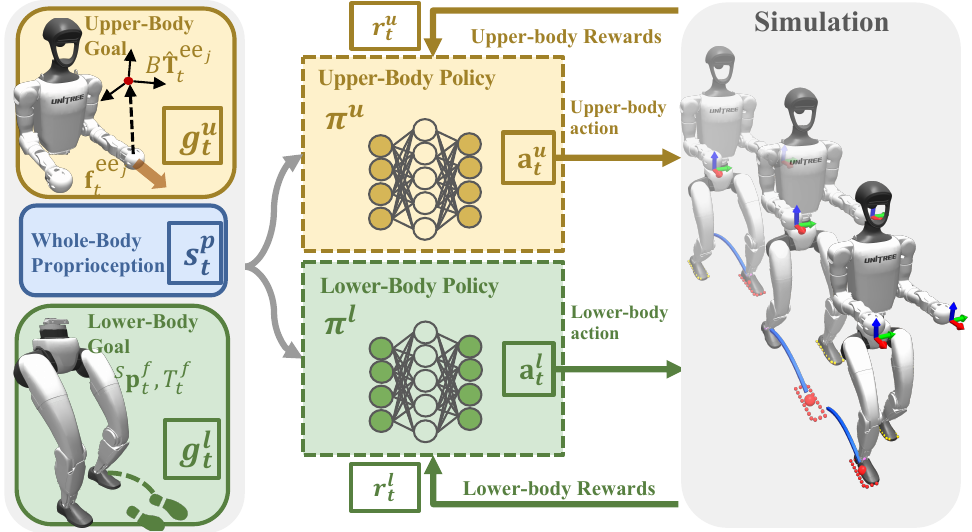}
    \caption{The low-level controller is composed of an upper- and a lower-body policy. The upper-body policy tracks the target end-effector pose and force, while the lower-body policy tracks footstep commands. The two policies are jointly trained with shared whole-body proprioception.}
    \label{fig:controller}
    \vspace{-0.1in}
\end{figure}
We then score each candidate action by its expected information gain (EIG) over these two outcomes:

\begin{equation}
\begin{aligned}
a_t
&=
\arg\max_{a\in\mathcal A} \;
\mathrm{EIG}_t(a),
\\
&=\arg\max_{a\in\mathcal A} \;
\left[p_t(a)\,\mathrm{IG}_t^{\mathrm{c}}(a)+(1-p_t(a))\,\mathrm{IG}_t^{\mathrm{f}}(a)\right],
\label{eq:eig_selection}
\end{aligned}
\end{equation}
where $\mathrm{IG}_t^{\mathrm{c}}(a)$ and $\mathrm{IG}_t^{\mathrm{f}}(a)$ denote the information gains of contact and contact-free outcomes, respectively. After $a_t$ is executed, the resulting tactile observation $o_t^{\mathrm{tac}}$ is fed back into the particle filter to update the belief. Action selection and execution are repeated on the updated belief until the maximum EIG falls below a threshold, at which point the robot proceeds to the next subgoal.

\begin{figure*}[t]
  \centering
  \includegraphics[width=\textwidth]{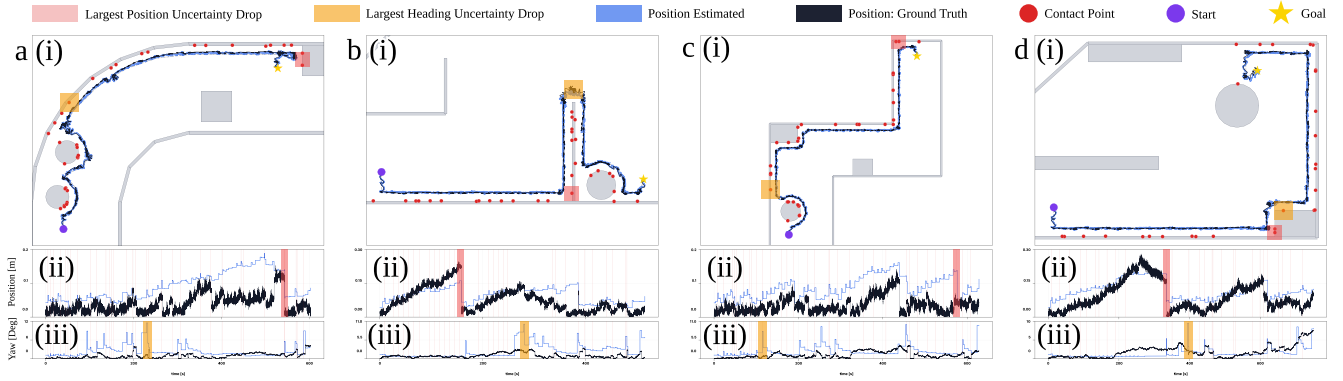}
  \caption{\method navigation performance in simulation. (a)--(d) correspond to four evaluated maps: (a) a curved corridor, (b) a partitioned room, (c) a zigzag corridor, and (d) a room with scattered obstacles. Each map has its own start (purple) and goal (yellow star). For each map, (i) shows the map layout with the estimated trajectory (blue), the ground-truth trajectory measured from the simulator (black), and contact points from probing (red), where each grid cell is $1\,\text{m}\times1\,\text{m}$; (ii), (iii) show position and yaw uncertainty (blue) and error (black) over time, respectively, with vertical red lines marking contact events. The red and yellow boxes highlight the largest decreases in both estimation error and uncertainty, which occur when the robot contacts non-planar wall surfaces.}
  \label{fig:maps}
\end{figure*}

\subsection{Low-level Controller}
\label{sec:controller}

To achieve navigation with frequent stop-and-go maneuvers while maintaining contact with the surroundings, we learn a whole-body control policy that tracks end-effector pose and force as well as footstep placement~\cite{gu2026refinedp}. As shown in Fig.~\ref{fig:controller}, similar to previous works~\cite{jang2025seecstableendeffectorcontrol}, we adopt a decoupled architecture in the controller. At time $t$, the upper-body goal is denoted as $\uppergoal = [\{{}^{B}\hat{\mathbf{T}}^{\text{ee}_j}_t,\; \mathbf{f}^{\text{ee}_j}_t,\; c^{\text{ee}_j}_t\}_{{j \in \{L,R\}}}]$, which is specified per end-effector $j$ and consists of 
the target end-effector pose 
${}^{B}\hat{\mathbf{T}}^{\text{ee}_j}_t \in \mathrm{SE}(3)$ in the robot base frame $\{B\}$, 
the target normal contact force $\mathbf{f}^{\text{ee}_j}_t \in \mathbb{R}_{\ge 0}$, 
and a binary contact indicator $c^{\text{ee}_j}_t \in \{0, 1\}$ for the force command. The lower-body goal $\lowergoal = [ {}^{S}\mathbf{p}^{f}_t, T^f_t, c^f_t]$ contains the target footstep 
${}^{S}\mathbf{p}^{f}_t \in \SE{2}$ 
expressed in the current stance-foot frame $\{S\}$, the footstep countdown $T^f_t \in \mathbb{R}$, and a binary foot-swing indicator $c^f_t \in \{0, 1\}$ of the current swing foot. The shared proprioception data $\proprio = [\mathbf{q}^r_{t-H:t}, \mathbf{\dot{q}}^r_{t-H:t}, \boldsymbol{\omega}^\text{r}_{t-H:t}, \mathbf{g}^\text{r}_{t-H:t}, \mathbf{a^\pi}_{t-H:t-1}]$ contains a history of $H$-timestep joint positions $\mathbf{q}^r_t \in \mathbb{R}^n$, joint velocities $\mathbf{\dot{q}}^r_t \in \mathbb{R}^n$, robot angular velocity $\boldsymbol{\omega}^\text{r}_t \in \mathbb{R}^3$, robot projected gravity $\mathbf{g}^\text{r}_t \in \mathbb{R}^3$, and whole-body actions $\mathbf{a}^\pi_t \in \mathbb{R}^n$. The upper-body policy $\pi^u$ takes the upper-body goal $\uppergoal$ and outputs upper-body actions $\upperaction \in \mathbb{R}^{n^u}$. The lower-body policy $\pi^l$ takes the locomotion goal $\lowergoal$ and outputs lower-body actions $\loweraction \in \mathbb{R}^{n^l}$. Formally, we learn an upper-body policy $\pi^u: \proprio \times \uppergoal \to \upperaction$ and a lower-body policy $\pi^l: \proprio \times \lowergoal \to \loweraction$. The parameters of the two policies, $\theta_l$ and $\theta_u$, are jointly updated with PPO~\cite{schulman2017proximalpolicyoptimizationalgorithms}, optimizing independent goal-conditioned objectives $\max_{\theta_{u}} \mathbb{E}[\sum_{t=1}^{T} \gamma^{t-1} r^{u}_t]$ and $\max_{\theta_{l}} \mathbb{E}[\sum_{t=1}^{T} \gamma^{t-1} r^{l}_t]$, where $r^{u}_t$ and $r^{l}_t$ denote the upper-body and lower-body rewards, respectively. The whole-body action $\mathbf{a^\pi}_t = [\upperaction, \loweraction] \in \mathbb{R}^n$ combines the upper-body and lower-body actions, and the final action sent to the low-level PD controller is $\hat{\mathbf{a}}_t = \mathbf{q}_\text{default} + \alpha \mathbf{a^\pi}_t$, where $\alpha$ is the action scale.

\subsubsection{Upper-body Policy Training} 
The goal of the upper-body policy includes accurate tracking of the target end-effector pose and force. A two-stage training curriculum is adopted: end-effector pose tracking at the first stage and force command tracking at the second stage. In the first stage, an upper-body reference joint position dataset is acquired by sampling end-effector poses ${}^{B}\hat{\mathbf{T}}^{\text{ee}_j}_t$ and solving inverse kinematics problems with posture regularization costs. The policy is trained to track both the reference joint positions and the corresponding end-effector poses.

In the second stage, similar to prior works~\cite{Portela_2024}, a virtual external force $\mathbf{F}^{\text{ext}_j}_t \in \mathbb{R}^3$ is applied to the end-effector in simulation. The force is produced by a spring-damper field acting along the probe direction $\mathbf{n}_t^{\text{ee}_j} \in \mathbb{R}^3$, taken as the $x$-axis of the end-effector frame $\{E_j\}$. Let $K_p, K_d \in \mathbb{R}$ be the stiffness and damping of the field, and $\mathbf{x}^{s}_t \in \mathbb{R}^3$ the setpoint of the field. The force magnitude is $f^{\text{n}}_t = \max\big(0,\; K_p \big(\mathbf{p}^{\text{ee}_j}_t - \mathbf{x}^{s}_t\big)^{\!\top}\mathbf{n}^{\text{ee}_j}_t + K_d \big(\dot{\mathbf{p}}^{\text{ee}_j}_t - \dot{\mathbf{x}}^{s}_t\big)^{\!\top}\mathbf{n}^{\text{ee}_j}_t\big) \in \mathbb{R}_{\ge 0}$,
where $\mathbf{p}^{\text{ee}_j}_t, \dot{\mathbf{p}}^{\text{ee}_j}_t \in \mathbb{R}^3$ are the current end-effector position and velocity. We place the setpoint so that holding the target pose requires exerting exactly the commanded normal force, which gives
$
\mathbf{x}^{s}_t = {}^{B}\hat{\mathbf{p}}^{\text{ee}_j}_t - \frac{f^{\text{ee}_j}_t}{K_p}\,\mathbf{n}^{\text{ee}_j}_t.
$
The applied force is the reaction of the field,
$
\mathbf{F}^{\text{ext}_j}_t = -f^{\text{n}}_t\,\mathbf{n}^{\text{ee}_j}_t.
$

\subsubsection{Lower-body Policy Training}

Our navigation task requires frequent stop-and-go transitions and precise adjustment of the robot's position, we therefore command the lower-body policy with footstep placements \cite{gu2026refinedp}. Given a double-support rest configuration and a footstep target $[{}^{S}\mathbf{p}^{f}_{t}, T^{f}_{t}, c^{f}_{t}]$, we generate a rest-to-rest step motion reference. A piecewise-constant Zero Moment Point (ZMP) reference is mapped to a center of mass (CoM) trajectory by the 3D linear inverted pendulum model (3D-LIPM) \cite{kajita2001lipm} under zero initial and terminal velocity.
The swing foot follows a cubic spline with zero lift-off and landing velocities and a fixed swing height, and the lower-body joint references are obtained by inverse kinematics. The dataset is built by sampling double-support configurations, footstep targets
${}^{S}\mathbf{p}^{f}_{t}$, and step durations $T^{f}_{t}$. The policy is then
trained to track the reference CoM, swing-foot, and lower-body joint
trajectories \cite{liu2024opt2skill}.

\section{Experiments}
\label{sec:experiments}

We evaluate \method in simulation and on a Unitree G1 humanoid
using the parameters in Table~\ref{tab:experiment_parameters}.
The experiments address three questions:

\textbf{Q1}: Does \method provide reliable and safely bounded state estimation throughout navigation across different maps? 

\textbf{Q2}: How accurately does the low-level controller adjust the robot's position throughout navigation?

\textbf{Q3}: Can \method perform reliably in real-world navigation tasks?

\begin{table}[!t]
\centering
\small
\setlength{\tabcolsep}{3pt}
\begin{tabular}{
@{}
>{\centering\arraybackslash}p{0.5\linewidth}
>{\centering\arraybackslash}p{0.46\linewidth}
@{}
}
\toprule
Parameter & Setting \\
\midrule
Controller Frequency & 200 / 50 Hz \\
Replanning Deviation Margin & $>0.1$ m  \\
Global Planner Grid Size  & 0.1 m \\
Minimal Wall Clearance  & 0.2 m \\
$|\Delta\mathbf x_t|_{\text{max}}$ & 0.40 m \\
$(\bar\sigma_p,\bar\sigma_\psi)$ & (0.18 m, $20^\circ$) \\
$N_p$ & 600 \\
Initial Pose Standard Deviation & 0.03 m/axis, $1.5^\circ$ yaw \\
Number of Probes per Subgoal& Max. 4 \\
EIG Threshold & 0.15 nats \\
$\sigma_c$ & 0.018 m \\
$\sigma_{\mathrm{control}}$ & 0.045 m \\
Trials per Map and Method & 20 (seeds 0 -- 19) \\
Simulation Time Limit & 1800 s \\
\bottomrule
\end{tabular}
\caption{Simulation and navigation parameters.}
\label{tab:experiment_parameters}
\vspace{-0.25in}
\end{table}

\subsection{Simulation Results}

\begin{table*}[t]
  \centering
  \footnotesize
  \setlength{\tabcolsep}{4pt}
  
  \newcommand{\pmv}[2]{$#1_{\pm#2}$}   
  \newcommand{\bpmv}[2]{$\boldsymbol{#1_{\pm#2}}$}
  
  \begin{tabular*}{\textwidth}{@{\extracolsep{\fill}}l*{12}{c}}
    \toprule
    \multirow{2}{*}{\textbf{Method}} &
    \multicolumn{3}{c}{\textbf{Map a}} &
    \multicolumn{3}{c}{\textbf{Map b}} &
    \multicolumn{3}{c}{\textbf{Map c}} &
    \multicolumn{3}{c}{\textbf{Map d}} \\
    \cmidrule(lr){2-4}\cmidrule(lr){5-7}\cmidrule(lr){8-10}\cmidrule(lr){11-13}
    & \textbf{PosErr} & \textbf{YawErr} & \textbf{SR}
    & \textbf{PosErr} & \textbf{YawErr} & \textbf{SR}
    & \textbf{PosErr} & \textbf{YawErr} & \textbf{SR}
    & \textbf{PosErr} & \textbf{YawErr} & \textbf{SR} \\
    & (cm) $\downarrow$ & (deg) $\downarrow$ & (\%) $\uparrow$
    & (cm) $\downarrow$ & (deg) $\downarrow$ & (\%) $\uparrow$
    & (cm) $\downarrow$ & (deg) $\downarrow$ & (\%) $\uparrow$
    & (cm) $\downarrow$ & (deg) $\downarrow$ & (\%) $\uparrow$ \\
    \midrule
    \textbf{\method}   & \bpmv{6.5}{2.5} & \bpmv{1.4}{0.4} & \textbf{95} & \bpmv{9.4}{3.6} & \pmv{2.8}{1.8} & \textbf{100} & \bpmv{5.9}{1.9} & \pmv{2.0}{1.3} & \textbf{100} & \pmv{16.8}{5.0} & \pmv{2.3}{1.6} & \textbf{85} \\
    
    Random-touch       & \pmv{9.6}{3.8} & \pmv{2.1}{1.4} & 90 & \pmv{11.4}{3.9} & \pmv{2.9}{1.4} & 70 & \pmv{8.7}{3.6} & \pmv{2.9}{1.7} & 75 & \pmv{16.6}{6.0} & \pmv{2.7}{1.8} & 75 \\
    
    Sweep-touch        & \pmv{11.7}{4.5} & \pmv{2.0}{1.1} & 50 & \pmv{11.2}{3.9} & \bpmv{2.5}{1.5} & 50 & \pmv{10.9}{4.2} & \pmv{2.4}{1.6} & 30 & \bpmv{13.7}{3.5} & \bpmv{1.32}{0.1} & 25 \\
    
    Odometry-only      & \pmv{23.4}{11.2} & \pmv{3.5}{2.9} & 35 & 36.5 & 4.0 & 5 & 13.2 & \textbf{0.6} & 5 & \pmv{28.5}{12.2} & \pmv{4.0}{4.6} & 45 \\
    \bottomrule
  \end{tabular*}
  \caption{MuJoCo simulation results. The navigation performance of \method and the baselines is evaluated across four maps. Errors are reported as mean$_{\pm\text{std}}$; entries without a standard deviation correspond to settings in which only a single trial succeeded.}
  \label{tab:sim-results}
  \vspace{-0.2in}
\end{table*}

All simulation experiments are conducted in MuJoCo. To address \textbf{Q1}, \method is evaluated on four layouts in
Fig.~\ref{fig:maps}: (a) a curved corridor, (b) a partitioned room,
(c) a zigzag corridor, and (d) a room with scattered obstacles. For each map, \method navigates from a known starting location to a goal location through active tactile probing.
Across all environments, \method maintains estimated trajectories closely following the ground-truth states measured from the simulator, indicating accurate localization during navigation. In Fig.~\ref{fig:maps}\textbf{(a)}--\textbf{(d)}\textbf{(i)}, the robot regularly makes contact with the environment. The contact points along the navigation paths are associated with uncertainty and error reduction. The largest uncertainty reductions in Fig.~\ref{fig:maps}\textbf{(a)}--\textbf{(d)}\textbf{(ii), (iii)} are closely related to the contact geometry; contacts with different surface normals around corners and pillars provide complementary position constraints, significantly reducing position uncertainty. On the other hand, flat-wall contacts only constrain position along the surface normal but can reduce heading uncertainty. Together, these results show that \method exploits active and informative tactile contacts to reduce localization uncertainty.

\begin{figure}[t]
    \centering
    \includegraphics[width=\linewidth]{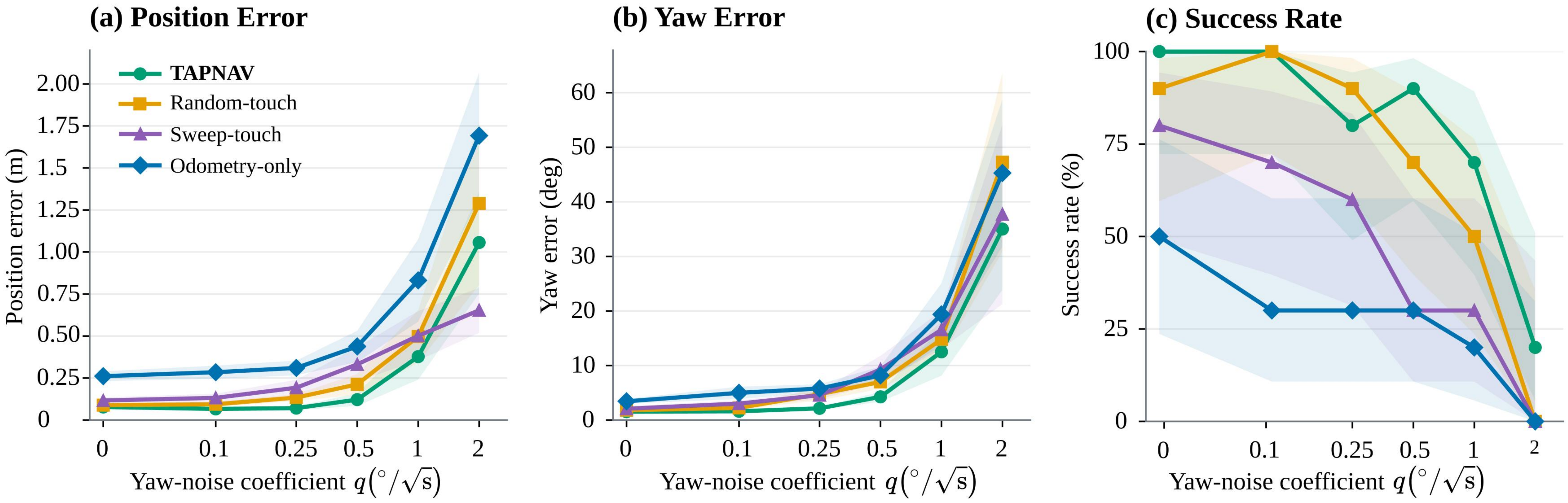}
    \caption{Yaw-noise sensitivity. (a) position error; (b) yaw error; (c) success rate. Bands and bars show mean $\pm$ standard error in (a)--(b) and 95\% confidence intervals in (c).}
    \label{fig:yaw_noise}
    \vspace{-0.2in}
\end{figure}

Table~\ref{tab:sim-results} reports an ablation study comparing the navigation performance of \method and the baselines across the four maps in Fig.~\ref{fig:maps}. The ablation study baselines include (i) Odometry-only, (ii) Sweep-touch, and (iii) Random-touch. Odometry-only disables tactile probing and relies entirely on the odometry and IMU updates for navigation, while Random-touch and Sweep-touch replace the EIG-driven action selection of Sec.~\ref{sec:eig_planner} with uniformly sampled probing actions and a grid sweep over the probing directions, respectively.
We report two metrics: the \textbf{pose estimation error}, given as the position and yaw errors between the torso pose estimated from the robot's belief and the ground truth, averaged over successful trials, and the \textbf{success rate}, defined as the fraction of trials where the robot reaches the goal region successfully. Each method is evaluated over $20$ trials per map, totaling $80$ trials across the four maps. The Odometry-only baseline achieves the lowest average success rate of $22.5\%$, highlighting the importance of active tactile sensing for vision-denied navigation. Compared with the two active tactile sensing baselines, \method achieves a higher average success rate of $95\%$, compared with $77.5\%$ for Random-touch and $38.8\%$ for Sweep-touch, while achieving a lower position error and comparable yaw error. 
These results support the effectiveness of \method in maintaining localization accuracy and completing navigation tasks.  To test whether these gains persist under increased heading uncertainty, we add yaw noise and compute errors over all completed trials including failures. Fig.~\ref{fig:yaw_noise} shows that \method achieves the lowest mean position and yaw errors for all IMU noise. It also retains the highest success rate, maintaining its navigation advantage even under additional heading noise.

\begin{figure}
    \centering
    \includegraphics[width=\linewidth]{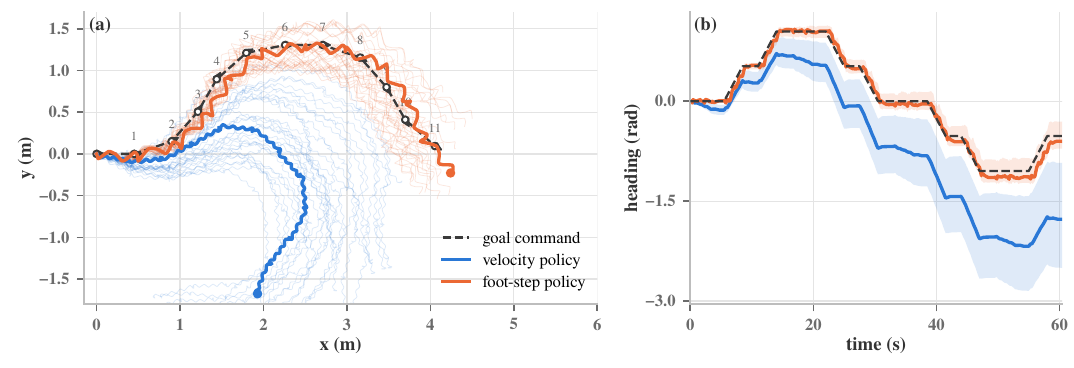}
    \caption{Simulated open-loop goal tracking with velocity- and footstep-tracking policies (30 trials per policy). (a) Executed trajectories, highlighting the trial with median final position error for each policy; (b) Actual and commanded heading over time.}
    \label{fig:controller_sim}
    \vspace{-0.3in}
\end{figure}

To answer \textbf{Q2}, we compare the performance of the \method low-level footstep controller with a velocity-tracking controller. The velocity-tracking controller adopts the same architecture as the \method footstep controller but has different goal for its lower-body policy, which is denoted as ${\lowergoal}' = [v^{r_x}_t, v^{r_y}_t, w^r_t] \in \SE{3}$ and contains robot linear velocity command in x,y directions and angular velocity command. Formally, the velocity-tracking controller consists of a upper-body policy ${\pi^u}': \proprio \times \uppergoal \to {\upperaction}$ and a lower-body policy ${\pi^l}': \proprio \times {\lowergoal}' \to {\loweraction}$.
The two controllers track a sequence of global planar torso targets under open-loop execution. Each target is given as a planar displacement $(\Delta x,\Delta y,\Delta\psi)$ from the previous target. Before execution, displacements are converted into lower-body goals for each controller, assuming the previous target was reached. Each target is given three
$1\,\mathrm{s}$ step periods, followed by a $2.5\,\mathrm{s}$ stand, so both
controllers follow the same nominal path: a straight line to the target at
constant speed, with the heading changing linearly. The velocity controller receives a constant velocity $(\Delta x,\Delta y)/3\,\mathrm{s}$ and yaw rate $\Delta\psi/3\,\mathrm{s}$, expressed in robot frame using the planned heading, and a zero command while standing. For the footstep controller, the displacement is divided into three evenly spaced waypoints. Each waypoint gives the landing pose of the next swing foot in current stance foot frame, shifted
sideways by half the nominal stance width, with the feet alternating, and a
final step brings robot back to nominal stance pose. Consecutive targets are separated by at most $0.5\,\mathrm{m}$, and the commanded velocity is limited to $0.2\,\mathrm{m/s}$ to mirror the target spacing and walking speeds of our navigation task. As shown in Fig.~\ref{fig:controller_sim}, heading errors accumulate across successive targets, causing the velocity-tracking controller to drift from the desired route. In contrast, the footstep controller reliably follows the target sequence and achieves smaller drift. 

\subsection{Real-World Experiments}

\begin{figure*}[t!]
    \centering
    \includegraphics[width=\linewidth]{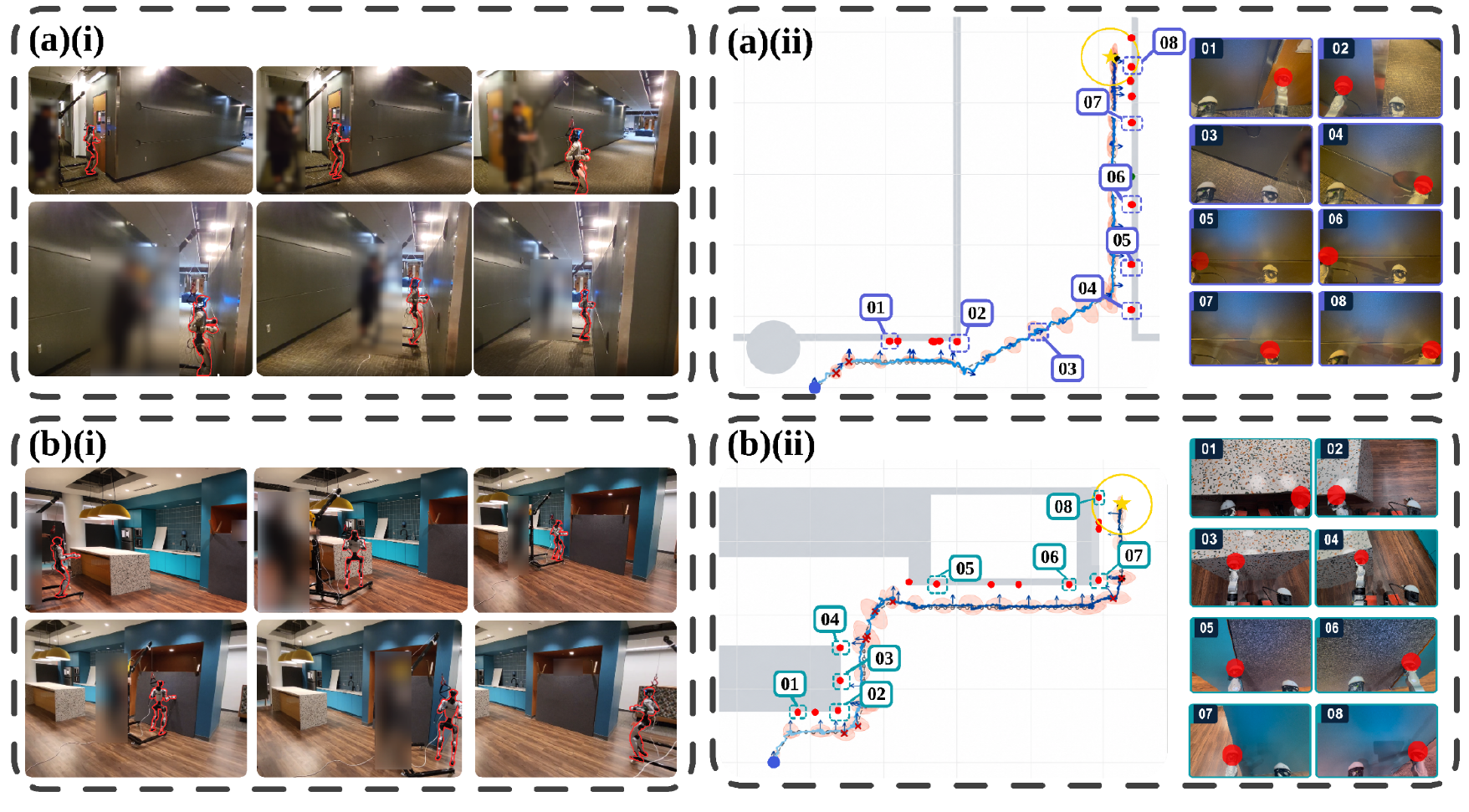}
    \caption{\method deployed on a Unitree G1 humanoid equipped with our tactile end-effectors, demonstrating real-world navigation capabilities in (a) a kitchen environment and (b) an office hallway. For each environment, (i) shows snapshots of the real-world deployment; (ii) shows the robot's path and active tactile probes in the robot's egocentric view.}
    \label{fig:realworld}
    \vspace{-0.2in}
\end{figure*}

To answer \textbf{Q3}, we evaluate \method on a Unitree G1 equipped with tactile end-effectors, as shown in Fig.~\ref{fig:hardware_design}. We conduct experiments in an office hallway and a kitchen environment to evaluate the integrated estimation, planning, and control pipeline under real-world sensing and control uncertainties. 
Fig.~\ref{fig:realworld} illustrates successful navigation trials in both environments. During execution, the robot moves around the counter, makes successive turns, and reaches the goal. Along the route, it actively probes different surfaces to reduce localization uncertainty, demonstrating effective tactile sensing across changes in contact geometry.

\section{Conclusion}
In this study, we presented \method, a vision-denied humanoid navigation framework integrating tactile-based active probing, belief-space localization, uncertainty-aware global planning, and whole-body control. Simulation and hardware experiments demonstrate goal-directed navigation using tactile feedback, with improved navigation success and lower pose error over baselines from the ablation study in simulation. A current limitation is that \method assumes a known static floor plan and approximates the particle belief as a unimodal Gaussian during planning, which we plan to relax by extending the framework to multimodal beliefs and partially known maps.

\section{Acknowledgment}
ChatGPT was used in Fig.~\ref{fig:teaser} to generate the uncertainty ellipses, in Fig.~\ref{fig:maps} for anti-aliasing using our original plot, and in Fig.~\ref{fig:realworld} for generating series number and individual facial mosaics, which we then manually added to the plots to cover the human faces.
\bibliographystyle{IEEEtran}
\bibliography{references}

\end{document}